\documentclass[10pt,twocolumn,letterpaper]{article}

\usepackage{cvpr}

\usepackage{xcolor, colortbl}
\definecolor{mono_orange}{RGB}{237,125,49}
\definecolor{mono_teal}{RGB}{143,206,211}

\definecolor{as_1}{RGB}{227,144,54}
\definecolor{as_2}{RGB}{195,121,79} 
\definecolor{as_3}{RGB}{167,127,100} 
\definecolor{as_4}{RGB}{142,132,119} 
\definecolor{as_5}{RGB}{117,138,138} 
\definecolor{as_6}{RGB}{98,142,154} 
\definecolor{as_7}{RGB}{56,150,185}
\definecolor{as_8}{RGB}{28,155,203}
\definecolor{as_9}{RGB}{0,160,215} 
\definecolor{as_10}{RGB}{0,170,230}

\newcommand{\model}{\textbf{%
\textcolor{as_1}{A}%
\textcolor{as_2}{d}%
\textcolor{as_3}{a}%
\textcolor{as_4}{S}%
\textcolor{as_5}{F}%
\textcolor{as_6}{o}%
\textcolor{as_7}{r}%
\textcolor{as_8}{m}%
\textcolor{as_9}{e}%
\textcolor{as_10}{r}%
}}

\usepackage{soul}
\newcommand{\hlorange}[1]{{\sethlcolor{mono_orange!20}\hl{#1}}}
\newcommand{\hlteal}[1]{{\sethlcolor{mono_teal!30}\hl{#1}}}

\definecolor{cvprblue}{rgb}{0.21,0.49,0.74}
\usepackage[pagebackref=false,breaklinks,colorlinks,allcolors=cvprblue]{hyperref}

\usepackage[accsupp]{axessibility}

\def\paperID{}
\def\confName{CVPR}
\def\confYear{2026}

\title{\vspace{-0.4cm}
\includegraphics[width=0.03\linewidth]{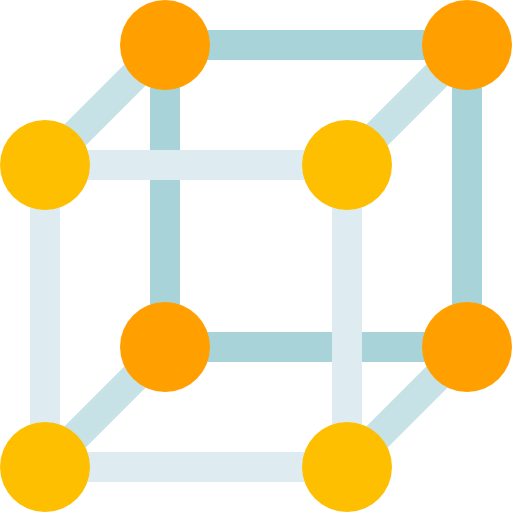}~\model: Adaptive Serialized Transformers\\for Monocular Semantic Scene Completion from Indoor Environments}

\author{
Xuzhi Wang$^1$ \quad
Xinran Wu$^1$ \quad
Song Wang$^2$ \quad
Lingdong Kong$^3$\thanks{Corresponding authors.}  \quad
Ziping Zhao$^1$\footnotemark[1] 
\\[1ex]
$^1$Tianjin Normal University \quad
$^2$Zhejiang University \quad
$^3$National University of Singapore \\
}

\begin{document}

\twocolumn[{
    \renewcommand\twocolumn[1][]{#1}
    \maketitle
    \begin{center}
    \centering
    \captionsetup{type=figure}
    \vspace{-0.7cm}
    \includegraphics[width=\textwidth]{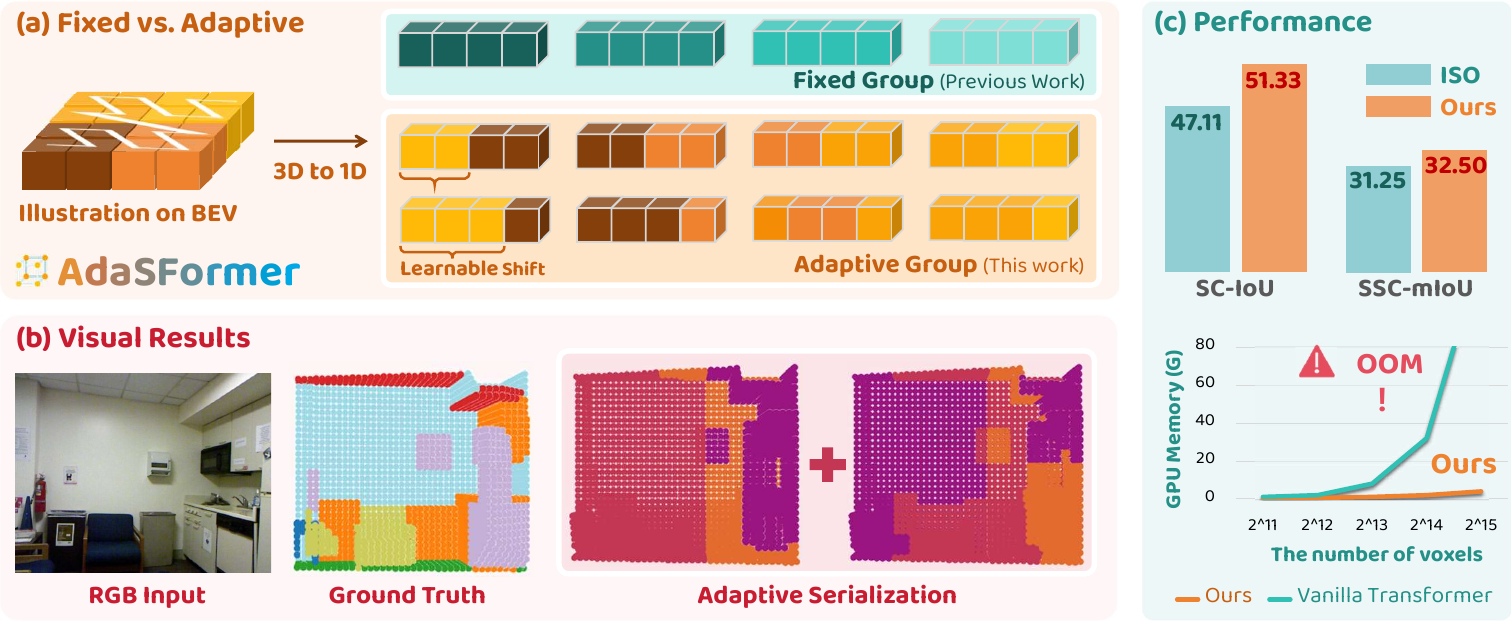}
    \vspace{-0.65cm}
    \caption{We introduce an \textbf{Adaptive Serialized Transformer (\model)} to obtain a global receptive field for indoor 3D scene understanding, enabling the model to handle complex spatial layouts and severe occlusions. \textbf{(1)} Existing indoor approaches rely on convolution-based designs due to the high memory cost of Transformers, which limits their global modeling capacity. \textbf{(2)} Existing serialization-based methods assign a fixed receptive field to each patch. In contrast, our method employs a learnable shift mechanism that adaptively adjusts the receptive fields and coordinate features across layers, achieving more flexible and effective spatial representation. It is worth noting that we employ bird’s-eye view (BEV) representations to simplify the 3D voxels, enabling clearer and more intuitive visualization.}
    \label{fig:teaser}
    \vspace{0.2cm}
    \end{center}
 }]
\maketitle

\maketitle
\renewcommand{\thefootnote}{*} 
\footnotetext{Corresponding authors.}
\renewcommand{\thefootnote}{\arabic{footnote}}

\begin{abstract}
Indoor monocular semantic scene completion (MSSC) is notably more challenging than its outdoor counterpart due to complex spatial layouts and severe occlusions. While transformers are well suited for modeling global dependencies, their high memory cost and difficulty in reconstructing fine-grained details have limited their use in indoor MSSC. To address these limitations, we introduce \textsl{\model}, a serialized transformer framework tailored for indoor MSSC. Our model features \textbf{three} key designs: \textbf{(1)} an Adaptive Serialized Transformer with learnable shifts that dynamically adjust receptive fields; \textbf{(2)} a Center-Relative Positional Encoding that captures spatial information richness; and \textbf{(3)} a Convolution-Modulated Layer Normalization that bridges heterogeneous representations between convolutional and transformer features. Extensive experiments on NYUv2 and Occ-ScanNet demonstrate that ASFormer achieves state-of-the-art performance. The code is publicly available at: \url{https://github.com/alanWXZ/AdaSFormer}.
\end{abstract} 
\vspace{-0.4cm}
\section{Introduction}
\label{sec:intro}

Monocular semantic scene completion (MSSC) has recently gained traction in the context of autonomous driving \cite{ssc_outdoor_MonoScene,wei2023surroundocc,ssc_outdoor_nuCraft,bian2025dynamiccity,xu2024superflow}, as it enables holistic 3D understanding of complex outdoor environments from a single image~\cite{ ssc_outdoor_VoxFomer, ssc_outdoor_HASSC}.
In contrast, indoor MSSC remains underexplored despite its fundamental importance to indoor scene perception, supporting applications in robot navigation, 3D reconstruction, and spatial understanding \cite{survey_3d_4d_world_models,chen2023clip2Scene,kong2025largead,kong2025multi,xu2025frnet,xie2025drivebench, feng2025reasonmap,xie2025robobev,xu2025lima}.

Indoor scenes differ substantially from outdoor settings \cite{kong2023laserMix,liu2023segment,liang2025lidarcrafter,liang2025perspective,kong2023robo3d,xu2025visual,xu2026superflow++}. Rather than featuring regular road layouts and open views, they contain intricate and multi-level spatial structures, including overlapping ceilings, walls, and furniture, with frequent occlusions and high object density \cite{chen2023towards,li2025seeground,wang2025nuc-net}. This complexity demands not only accurate local geometry but also strong global contextual reasoning to infer the geometry and semantics of occluded regions \cite{ssc_MonoMRN}.

Most existing approaches heavily rely on convolutional architectures~\cite{ssc_MonoMRN,ssc_NDC-scene,ssc_ISO,SSCNet}. However, their inherently limited receptive fields prevent them from modeling long-range dependencies effectively, while enlarging kernels in 3D incurs prohibitive cubic computational cost \cite{ding2022scaling, chen2023largekernel3d}. Early convolutional layers propagate information solely within local neighborhoods, while deeper layers operate at reduced resolutions, leading to spatial information loss and degraded completion quality \cite{lu2023link, feng2024lsk3dnet, zhang2025scaling, li2024place3d}. 

Modeling long-range dependencies is therefore essential, making transformer architectures a natural choice \cite{vaswani2017attention, Swin_Transformer, han2022survey}. Yet, applying transformers directly to dense 3D volumes remains computationally expensive and memory-intensive \cite{mao2021voxel, he2022voxel, li2022unifying, ssc_outdoor_VoxFomer, ser_wu2024point}, especially for high-resolution indoor scenes \cite{dataset_nyu, ssc_ISO}.

To address these challenges, we design \textsl{\model}, a hybrid framework that unifies convolutional and transformer modules, each serving complementary roles. The transformer operates on voxels containing valid features to capture global context, while the convolution focuses on local feature propagation and refinement. Our framework introduces \textbf{three} core designs:
\textbf{(1)} \textit{Adaptive Serialized Attention} (ASA), which employs learnable shifts to dynamically adjust receptive fields and coordinate features across layers;
\textbf{(2)} \textit{Center-Relative Positional Encoding (CRPE)}, which encodes spatial information richness relative to the scene center; and
\textbf{(3)} \textit{Convolution-Modulated Layer Normalization} (CMLN), which harmonizes feature statistics between convolutional and transformer representations.

\begin{figure*}[t]
    \centering
    \includegraphics[width=0.89\textwidth]{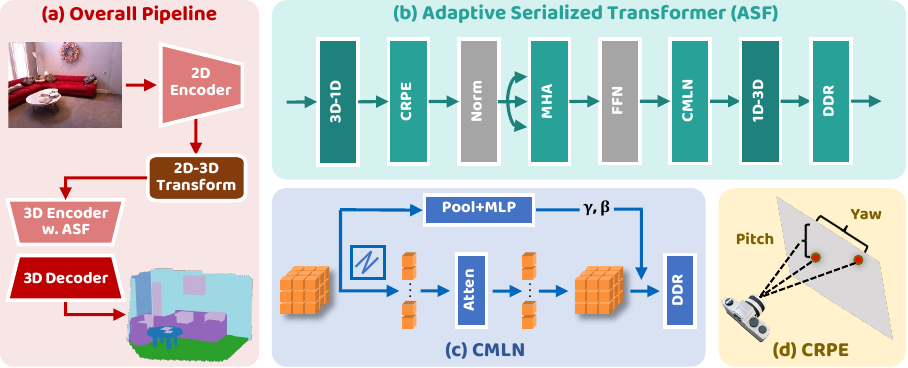}
    \vspace{-0.2cm}
    \caption{\textbf{\model~Framework}. (a) Overall network architecture: It mainly consists of a 2D encoder, a depth estimation network, a 2D-to-3D projection module, and a 3D network. (b) Adaptive Serialized Transformer (ASF): This is the core of our method. A set of ASF blocks forms the encoder of the 3D network. The 3D voxels are adaptively converted into 1D patches, which are then augmented with center-relative positional encoding. These patches are processed through standard self-attention and feed-forward layers. Through the proposed Convolution-Modulated Layer Normalization (CMLN), the heterogeneous features from the transformer and convolutional layers are bridged. The resulting features are then fed into a DDR block for scene completion. (c) Convolution-Modulated Layer Normalization (CMLN): Modulates features to integrate transformer and convolutional representations effectively. (d) Center-Relative Positional Encoding: Enhances the richness of input information by modeling the relative spatial relationships of patches.} 
    \label{fig:framework}
\end{figure*}

As illustrated in \cref{fig:framework}, given a single RGB image, a 2D encoder first extracts features and estimates depth, which are projected into 3D space via surface projection~\cite{SATNet}.
Within the 3D encoder, alternating transformer and convolutional modules jointly capture global context and complete 3D geometry, followed by a lightweight decoder that integrates multi-scale features to produce final semantic scene completion (SSC) results.

Our main contributions are summarized as follows:
\begin{itemize}
\item We propose \textsl{\model} to capture long-range dependencies while maintaining efficiency through complementary feature propagation, substantially reducing both memory footprint and computational overhead.
\item We introduce three novel components, $^1$\textit{Adaptive Serialized Attention}, $^2$\textit{Center-Relative Positional Encoding}, and $^3$\textit{Convolution-Modulated Layer Normalization}, which collectively enhance receptive-field adaptability, spatial reasoning, and cross-module feature alignment.
\item Extensive experiments on the NYUv2 and OccScanNet datasets demonstrate that our approach achieves state-of-the-art performance against popular MSSC methods.
\end{itemize}

\section{Related Work}
\label{sec:related_work}

\noindent\textbf{Monocular Semantic Scene Completion (MSSC).}
Scene completion tasks predicts both voxel occupancy and semantic labels, providing holistic 3D scene understanding~\cite{li2025_3eed,FFNet,wang2025nuc-net, wang2025forging, wang2025pointlora, liang2025perspective}. The pioneering SSCNet~\cite{ssc_outdoor_MonoOcc} introduced indoor SSC using depth maps, later extended to RGB-D inputs~\cite{FFNet,AICNet,SATNet,DDRNet,Sketch-Net} and outdoor perception~\cite{ssc_outdoor_JS3CNet,ssc_outdoor_SCPNet,ssc_pami_lidar,liu2026veila,liu2026lalalidar,xu2025limoe,zhu2025spiral}. Recent monocular SSC approaches~\cite{ssc_outdoor_HTCL,ssc_outdoor_OccFormer,ssc_outdoor_Symphonize,ssc_outdoor_Occworld,ssc_outdoor_veon,ssc_outdoor_points,ssc_outdoor_viewformer,ssc_outdoor_nuCraft,ssc_outdoor_VoxFomer,ssc_outdoor_LowRankOcc, wang2025reliable, wang2024label} further advance the task by relying solely on RGB images. For indoor scenarios, where lightweight deployment is critical, several representative works have emerged: MonoScene~\cite{ssc_outdoor_MonoScene} bridges 2D-3D features via Line-of-Sight Projection; NDC-Scene~\cite{ssc_NDC-scene} introduces normalized device coordinates to reduce depth ambiguity; ISO~\cite{ssc_ISO} enhances voxel learning with dual projection; and MonoMRN~\cite{ssc_MonoMRN} adopts a masked recurrent refinement design. Most indoor MSSC methods are CNN-based, achieving efficiency but limited global reasoning. In contrast, our approach employs an adaptive serialized transformer to enlarge receptive fields while maintaining low computational cost.

\noindent\textbf{Single-View 3D Reconstruction.} 
Recovering 3D geometry from a single RGB image has long been studied for both objects and scenes \cite{liu2024multi,jaritz2020xMUDA}. Early works reconstruct individual objects using explicit~\cite{3D_RGB_2018icml,3D_RGB_3D-R2N2,3D_RGB_point_set} or implicit~\cite{3D_RGB_DeepSDF,3D_RGB_convolution_occupancy,3D_RGB_occupancy_network} representations, while more recent efforts address full scene reconstruction~\cite{3DR_zhao2025depr,3DR_chen2024comboverse,3DR_liu2023zero}. Holistic reconstruction methods~\cite{3D_RGB_panoptic,3D_RGB_BUOL,3DR_yan2023psdr,3DR_liu2022towards,3DR_zhang2021holistic} recover the scene as a whole, often guided by depth estimation~\cite{ssc_outdoor_VoxFomer,ssc_ISO,ssc_outdoor_MonoScene} and processed through encoder–decoder architectures. Compositional approaches~\cite{3DR_zhou2024zero,3DR_liu2023zero,3DR_gao2024diffcad,3DR_gumeli2022roca} reconstruct objects separately before assembling them via layout and pose estimation, as seen in DepR~\cite{3DR_zhao2025depr} and ComboVerse~\cite{3DR_chen2024comboverse}. These reconstruction paradigms provide geometric priors and structural insights closely related to semantic scene completion.

\noindent\textbf{Serialization-Based Approaches.} 
The recent serialization-based transformers~\cite{ser_chen2022efficient,ser_wang2023octformer,ser_liu2023flatformer,ser_wu2024point} convert irregular 3D data into ordered sequences to enable efficient attention while preserving spatial locality. Representative works include OctFormer~\cite{ser_wang2023octformer}, which partitions point clouds via sorted octree keys; FlatFormer~\cite{ser_liu2023flatformer}, which reduces padding through flattened window grouping; and PTv3~\cite{ser_wu2024point}, which applies shuffle-based ordering with conditional positional encoding. Although designed mainly for 3D segmentation and detection, these methods demonstrate the efficiency of serialized attention for large-scale 3D modeling. Our work extends this idea to semantic scene completion through adaptive serialization with learnable shifts and center-relative positional encoding, improving long-range reasoning under sparse 3D layouts.
\section{Methodology}
\label{sec:method}

In this section, we first provide an overview of our method. Next, we give a brief preliminary to voxel serialization (\cref{Voxel Serialization}). 
Following this, we delve into the AdaSFormer network, including its three proposed key designs: Adaptive Serialized Attention (\cref{sec: Adaptive Serialized Attention}), Center-Relative Positional Encoding (\cref{sec: Center-Relative Positional Encoding}), and the Conv-Modulated Layer Normalization (\cref{sec: Conv-Modulated Layer Normalization}). Lastly, we introduce the loss function for model training (\cref{sec: loss function}).

\noindent\textbf{Problem Formulation.}  
Given a single-view RGB image $I_{\mathrm{RGB}}$ that captures only a partial 3D scene, monocular semantic scene completion aims to predict the volumetric occupancy of the entire scene. Each voxel is assigned a semantic label $C_i$ with $i \in [0, 1, \dots, N]$, where $N$ is the number of object categories and $C_0$ denotes empty space.

\noindent\textbf{Overall Network Architecture.} 
From a single RGB image, 2D features are first extracted using EfficientNet~\cite{tan2019efficientnet}, and the corresponding depth map is estimated using an off-the-shelf method~\cite{depthanything_v2}. Leveraging the estimated depth together with the camera’s intrinsic and extrinsic parameters, the 2D features are projected into 3D space. The resulting 3D representations are subsequently processed by a transformer-based encoder, consisting of a series of ASFormer blocks, followed by a decoder to produce the final Semantic Scene Completion (SSC) outputs. We refrain from employing a transformer-based architecture at the decoding stage, as the number of non-empty voxels increases significantly. In this setting, convolutional decoding offers higher efficiency than transformer-based decoding, effectively reducing computational overhead.

\subsection{Preliminary of Voxel Serialization}
\label{Voxel Serialization}
In this section, we present preliminaries of  serialization-based transformers.
We introduce a voxel serialization-based method that enables 3D networks to capture a sufficiently large and high-resolution receptive field in the early stage while keeping memory consumption manageable, which is crucial for scene completion tasks.

\noindent\textbf{Serialization.}  Serialization transforms irregular or high-dimensional spatial data into a structured sequential format suitable for attention-based modeling as shown in \cref{fig:zshift} (a). Specifically, given an input 3D voxel
$X \in \mathbb{R}^{C \times D \times H \times W}$, we employ a space-filling curve (\textit{e.g.}, Z curve and Hilbert curve) to linearize the volumetric representation along a continuous path. This preserves the local spatial proximity of neighboring elements, allowing the attention mechanism to operate efficiently while maintaining spatial correlations. After serialization, the data is represented as a sequence of tokens 
$S = [s_1, s_2, \dots, s_N]$, where $N = D \times H \times W$.

\noindent\textbf{Local Grouping.} 
The core purpose of serialization is to enable more effective local grouping. Since the computational complexity of the attention mechanism grows quadratically with the input sequence length (\(O(N^2)\)), reasonable data grouping is crucial for processing large-scale 3D data. Local grouping divides the serialized sequence into fixed-length groups, each containing \(G\) consecutive tokens. Attention is computed only within each group, which captures local dependencies while significantly reducing computational cost. Formally, the sequence is divided into \(M = \lceil N / G \rceil\) groups, where $\lceil \cdot \rceil$ denotes the ceiling operation (i.e., rounding up), ensuring that all tokens are included: 
\begin{equation}
    S = [S_1, S_2, \dots, S_M], \quad S_i \in \mathbb{R}^{G \times C}.
\end{equation}
Under this grouping, the computational complexity per group is \(O(G^2)\), leading to a total complexity of \(O(M \cdot G^2) = O(N \cdot G)\), which is much lower than the full attention complexity \(O(N^2)\).

Compared with conventional window-attention~\cite{Swin_Transformer} based grouping methods, serialization-based grouping for 3D data has the following advantages: 1) Avoids voxel padding overhead: 3D point clouds are often sparse, and in window-based grouping, the number of voxels in each window usually varies. Padding is required to unify the length, which wastes significant computational resources. In contrast, serialization-based grouping can evenly divide the non-empty voxels without additional padding, reducing redundant computation. 2) Larger effective receptive field: Window-based grouping includes empty voxels, limiting the local receptive field. Given the same number of tokens per group, serialization-based grouping removes empty voxels, allowing attention to operate over a wider range of valid spatial regions and capturing richer local contextual information.

\subsection{AdaSFormer: Towards Efficient MSSC}
Several stacked AdaSFormer blocks serve as the 3D network encoder, modeling global dependencies with moderate memory consumption. 
\subsubsection{Adaptive Serialized Attention (ASA)}
\label{sec: Adaptive Serialized Attention}
\begin{figure*}[t]
    \centering
    \includegraphics[width=\textwidth]{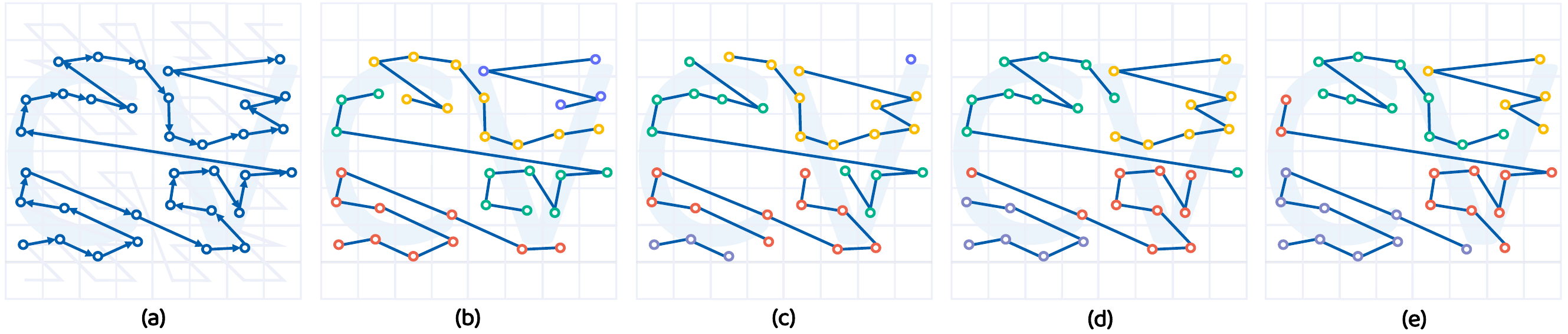}
    \vspace{-0.6cm}
    \caption{(a) illustrates the Z-order serialization of a 2D image, as well as that of a sparse 2D image, where the letters “C” and “V” in the figure indicate the locations of objects in the spatial domain. (b)-(e) illustration of Z-order serialization with shifts of 0, 3, 6, and 9 (from left to right). Different shift settings significantly affect the coverage and overlap between adjacent patches, especially for large patch sizes.}
    \label{fig:zshift}
\end{figure*}

Although the serialized transformer reduces GPU memory consumption by serializing patches, the obtained patches remain fixed once a specific serialization strategy is chosen. However, the grouping scheme also greatly affects the receptive field range. As illustrated in Figure~\ref{fig:zshift}, different starting points can significantly alter the receptive field: in some cases, the partitioned voxel range may fully cover a single object, thereby capturing its overall structure more effectively; in other cases, the voxel range may simultaneously include multiple objects and their spatial relationships. Such variations directly influence the model’s representational capacity at both local and global levels, as well as its ability to capture semantic correlations.

Therefore, obtaining an appropriate shift is crucial for serialized attention. However, existing serialized attention methods do not consider the influence of patch shifts. For window-based attention mechanisms, although prior works such as the Swin Transformer~\cite{Swin_Transformer} introduce the concept of window shifting, the shift size remains fixed and non-learnable. Moreover, since the Swin Transformer operates on 2D patches, its potential shift range is inherently constrained by the spatial window size. In contrast, serialized attention operates along a 1D voxel sequence, offering a broader and more flexible space for shift adjustment. This property makes it particularly suitable for incorporating learnable, data-driven shifts that can dynamically adapt the serialization starting point to better capture object structures and their spatial relationships.

Assuming a patch size of $P$, we introduce $K$ learnable parameters to model the spatial offsets within each voxel group after serialization. Each parameter corresponds to a discrete shift value uniformly spaced by $P / K$; that is, the $k$-th parameter represents a shift of $k \cdot (P / K)$. During training, these learnable offsets are optimized to capture the most informative spatial alignment of serialized voxels with respect to object boundaries or scene layouts, thereby improving the contextual consistency of the representation.

\noindent \textbf{Straight-Through Gumbel-Softmax.} By leveraging the differentiable relaxation provided by Gumbel-Softmax, the start point can be trained end-to-end. During training, we compute a straight-through estimator to allow gradient flow while maintaining discrete selections. Let $\mathbf{l}$ denote the logits and $\tau$ the temperature. The soft Gumbel-Softmax output $\mathbf{y}_{\mathrm{soft}}$ and the hard one-hot output $\mathbf{y}_{\mathrm{hard}}$ are computed as:

\begin{equation}
\mathbf{y}_{\mathrm{soft}} = \mathrm{softmax}\Big(\frac{\mathbf{l} + \mathbf{g}}{\tau}\Big), 
\quad \mathbf{g} \sim \mathrm{Gumbel}(0,1),
\end{equation}

\begin{equation}
\mathbf{y}_{\mathrm{hard}} = \mathrm{one\_hot}\Big(\arg\max_i \mathbf{y}_{\mathrm{soft},i}\Big).
\end{equation}

The straight-through (ST) Gumbel-Softmax output is then given by:
\begin{equation}
\mathbf{y}_{\mathrm{ST}} = \mathbf{y}_{\mathrm{hard}} + \mathbf{y}_{\mathrm{soft}} - \mathbf{y}_{\mathrm{soft}}.\mathrm{detach}().
\end{equation}
In this formulation, the forward pass uses the discrete $\mathbf{y}_{\mathrm{hard}}$, while the gradient during backpropagation flows through $\mathbf{y}_{\mathrm{soft}}$, enabling differentiable learning with discrete selections.

\noindent \textbf{Temperature Annealing Strategy.} We design a temperature annealing strategy to allow the network to better learn discrete selections. Specifically, the temperature $\tau$ in the Gumbel-Softmax is gradually decreased during training to control the discreteness of the sampling. At epoch $t$, $\tau$ is updated as:
\begin{equation}
\tau_t = \max\Big(\tau_{\min}, \, \tau_{\mathrm{init}} \cdot \exp(-\alpha t)\Big),
\end{equation}
where $\tau_{\mathrm{init}}$ is the initial temperature, $\tau_{\min}$ is the minimum temperature, and $\alpha$ is the annealing rate.

\subsubsection{Center-Relative Positional Encoding (CRPE)} 
\label{sec: Center-Relative Positional Encoding}
Since our algorithm adopts a hybrid architecture of CNN and Transformer, where the CNN effectively captures local positional information, the model has a reduced reliance on traditional positional encoding. Inspired by work~\cite{bae2025three}, which demonstrates that the spatial distribution of input information in 3D space has a significant impact on SSC performance. Based on this observation, we design a positional encoding scheme to model this information richness. Specifically, we compute the yaw and pitch angles between each point and the scene center to describe their relative spatial relationships. This encoding enables the model to perceive the spatial relations between different regions and the scene center, allowing it to identify areas that are more likely to contain key structural or semantic information, and thereby allocate attention and feature weights more effectively.

We compute the scene center $\mathbf{c}$ as the mean coordinate of all occupied voxels:
\begin{equation}
\mathbf{c} = 
\frac{\sum_{i=1}^{M} \mathbf{p}_i \cdot \mathbf{1}(v_i > 0)}
     {\sum_{i=1}^{M} \mathbf{1}(v_i > 0)},
\end{equation}
where $\mathbf{p}_i = (x_i, y_i, z_i)$ denotes the center coordinate of the $i$-th voxel, $v_i$ is its occupancy value, $\mathbf{1}(v_i > 0)$ is an indicator function that equals 1 if the voxel is occupied and 0 otherwise,  and $\mathbf{c}$ represents the mean coordinate of all occupied voxels, \textit{i.e.}, the scene center.

The yaw angle (horizontal rotation around the vertical axis) for each voxel and for the grid center is defined as:
\begin{equation}
\theta^i = \operatorname{atan2}(d^i_x, d^i_y),
\end{equation}
\begin{equation}
\theta^c = \operatorname{atan2}(d^c_x, d^c_y),
\end{equation}
where \(d^i_x\) and \(d^i_y\) are the \(x\) and \(y\) components of the vector \(\mathbf{d}^i\). The yaw difference between voxel \(i\) and the center is then given by:
\begin{equation}
\Delta \theta^i = \theta^i - \theta^c.
\end{equation}

Similarly, the pitch angle (vertical inclination) for each voxel and for the grid center are defined as:
\begin{equation}
\phi^i = \operatorname{atan2}\Big(d^i_z, \sqrt{(d^i_x)^2 + (d^i_y)^2}\Big),
\end{equation}
\begin{equation}
\phi^c = \operatorname{atan2}\Big(d^c_z, \sqrt{(d^c_x)^2 + (d^c_y)^2}\Big),
\end{equation}
and the pitch difference is computed as:
\begin{equation}
\Delta \phi^i = \phi^i - \phi^c.
\end{equation}

After obtaining $\Delta \phi$ and $\Delta \theta$, we concatenate them into a single feature vector, which is then processed by a multi-layer perceptron (MLP). The resulting output is used as a bias added to the serialized attention.

\subsubsection{Convolution-Modulated Layer Norm (CMLN)}
\label{sec: Conv-Modulated Layer Normalization}

Since transformers and convolutional networks extract fundamentally different types of features, directly interleaving these modules can cause learning difficulties and limit performance gains. To address this, we propose Conv-Modulated Layer Normalization motivated by~\cite{Peebles_2023_ICCV}. Specifically, after voxel features $X_{\text{voxel}}$ are serialized and processed by Serialized Attention, they are modulated through Conv-Modulated Layer Normalization before being fed into the DDR module, a lightweight component designed for semantic scene completion (SSC) refinement.

The Conv-Modulated Layer Normalization can be defined as follows:

\begin{equation}
\text{CMLN}(h_i \mid X_{\text{voxel}}) = \gamma(X_{\text{voxel}}) \odot \frac{h_i - \mu_i}{\sigma_i} + \beta(X_{\text{voxel}}),
\end{equation}
where $h_i$ denotes the features to be normalized, $\mu_i$ and $\sigma_i$ are the mean and standard deviation computed over the feature dimension, and $\gamma(X_{\text{voxel}})$, $\beta(X_{\text{voxel}})$ are functions (\textit{e.g.}, small MLPs) that produce the normalization parameters from the voxel features. This design allows the network to adaptively modulate feature statistics based on the input context.

\definecolor{ceiling}{RGB}{214, 38, 40}
\definecolor{floor}{RGB}{43, 160, 43}
\definecolor{wall}{RGB}{158, 216, 229}
\definecolor{window}{RGB}{114, 158, 206}
\definecolor{chair}{RGB}{204, 204, 91}
\definecolor{bed}{RGB}{255, 186, 119}
\definecolor{sofa}{RGB}{147, 102, 188}
\definecolor{table}{RGB}{30, 119, 181}
\definecolor{tvs}{RGB}{188, 188, 33}
\definecolor{furniture}{RGB}{255, 127, 12}
\definecolor{objects}{RGB}{196, 175, 214}

\begin{table*}
    \centering
    \caption{
        \textbf{Comparisons among state-of-the-art MSSC methods} on the \textbf{NYUv2} dataset \cite{dataset_nyu}. The symbol $\dag$ denotes the results presented by~\cite{ssc_outdoor_MonoScene}. The \hlorange{Best} and \hlteal{2nd Best} scores from each metric are highlighted in \hlorange{Orange} and \hlteal{Teal}, respectively.
    }
    \vspace{-0.2cm}
    \resizebox{\linewidth}{!}{
    \begin{tabular}{r|r|c|ccccccccccc|c}
    \toprule
    Method & Venue & \rotatebox{90}{SC IoU (\%)} & \rotatebox{90}{\textcolor{ceiling}{$\blacksquare$}~Ceiling} & \rotatebox{90}{\textcolor{floor}{$\blacksquare$}~Floor} & \rotatebox{90}{\textcolor{wall}{$\blacksquare$}~Wall} & \rotatebox{90}{\textcolor{window}{$\blacksquare$}~window} & \rotatebox{90}{\textcolor{chair}{$\blacksquare$}~chair} & \rotatebox{90}{\textcolor{bed}{$\blacksquare$}~bed} & \rotatebox{90}{\textcolor{sofa}{$\blacksquare$}~sofa} & \rotatebox{90}{\textcolor{table}{$\blacksquare$}~table} & \rotatebox{90}{\textcolor{tvs}{$\blacksquare$}~tvs} & \rotatebox{90}{\textcolor{furniture}{$\blacksquare$}~furniture} & \rotatebox{90}{\textcolor{objects}{$\blacksquare$}~objects} & \rotatebox{90}{SSC mIoU (\%)}
    \\
    \midrule\midrule
    LMSCNet$^\dag$ \cite{ssc_outdoor_LMSCNet} & 3DV'20 & $33.93$ & $4.49$ & $88.41$ & $4.63$ & $0.25$ & $3.94$ & $32.03$ & $15.44$ & $6.57$ & $0.02$ & $14.51$ & $4.39$ & $15.88$
    \\
    AICNet$^\dag$ \cite{AICNet} & CVPR'20 & $30.03$ & $7.58$ & $82.97$ & $9.15$ & $0.05$ & $6.93$ & $35.87$ & $22.92$ & $11.11$ & $0.71$ & $15.90$ & $6.45$ & $18.15$
    \\
    3DSketch$^\dag$ \cite{Sketch-Net} & CVPR'20 & $38.64$ & $8.53$ & $90.45$ & $9.94$ & $5.67$ & $10.64$ & $42.29$ & $29.21$ & $13.88$ & $9.38$ & $23.83$ & $8.19$ & $22.91$
    \\
    MonoScene \cite{ssc_outdoor_MonoScene} & CVPR'22 & $42.51$ & $8.89$ & \hlteal{$93.50$} & $12.06$ & $12.57$ & $13.72$ & {$48.19$} & {$36.11$} & $15.13$ & $15.22$ & $27.96$ & $12.94$ & $26.94$ 
    \\
    NDC-Scene \cite{ssc_NDC-scene} & ICCV'23 & {$44.17$} & {$12.02$} & \hlorange{$93.51$} & {$13.11$} & $13.77$ & {$15.83$} & {$49.57$} & {$39.87$} & {$17.17$} & {$24.57$} & {$31.00$} & {$14.96$} & {$29.03$}
    \\
    ISO \cite{ssc_ISO} & ECCV'24 & {$47.11$} & {$14.21$} & {$93.47$} & {$15.89$} & \hlteal {$15.14$} & \hlteal{$18.35$} & \hlteal{$50.01$} & \hlteal{$40.82$} & \hlteal{$18.25$} & \hlorange{$25.90$} & \hlteal{$34.08$} & {$17.67$} & \hlteal{$31.25$}
    \\ 
    MonoMRN~\cite{ssc_MonoMRN} & ICCV'25 & \hlorange{$53.16$} & \hlorange{$26.80$} & $92.02$ & \hlorange{$19.39$} & \hlorange{$18.50$} & {$17.66$} & $44.60$ & $31.02$ & \hlorange{$19.60$} & {$17.22$} & {$32.90$} & \hlteal{$18.31$} & {$30.73$}
    \\
    GenFuSE~\cite{selvakumar2025fake} & arXiv'25 & {$46.30$} & {$ 9.80$} & $ 92.80$ & {$13.80$} & {$13.70$} & {$ 17.60$} & $ 48.60$ & $ 39.70$ & {$ 17.60$} & \hlteal{$24.70$} & {$ 33.40$} & {$ 17.20$} & {$29.90$}
    \\
    \midrule
    \textbf{\model} & \textbf{Ours} & \hlteal{$51.33$} & \hlteal{$16.18$} & {$92.72$} & \hlteal{$16.96$} & $14.10$ & \hlorange{$23.93$} & \hlorange{$53.89$} & \hlorange{$44.52$} & {$16.65$} & {$23.85$} & \hlorange{$35.97$} & \hlorange{$18.67$} & \hlorange{$32.50$}
    \\ 
    \bottomrule
    \end{tabular}}
    \label{tab:NYUv2}
\end{table*}

After the Convolution-Modulated Layer Normalization, the features are fed into the DDR block~\cite{DDRNet}, a lightweight 3D convolutional module designed for semantic scene completion. The Adaptive Serialized Attention captures global contextual dependencies, while the convolution propagates information to spatially neighboring regions, effectively enhancing local geometric representations. Moreover, by incorporating convolutional operations, the Transformer avoids computing attention over all voxels, thereby reducing computational overhead.

\subsubsection{Loss Function}
\label{sec: loss function}
Our method is trained end-to-end from scratch by applying cross-entropy loss, geometric affinity loss, and semantic affinity loss~\cite{ssc_outdoor_MonoScene}: 
\begin{align}
    \mathcal{L}_\mathrm{total} = \mathcal{L}_\mathrm{ce} +  \mathcal{L}_\mathrm{scal}^{\mathrm{sem}} + \mathcal{L}_\mathrm{scal}^{\mathrm{geo}},
\end{align}
where $\mathcal{L}_\mathrm{scal}$ denotes the Scene-Class Affinity Loss proposed by MonoScene~\cite{ssc_outdoor_MonoScene}.

\section{Experiments}
We conduct experiments on the widely used NYUv2 dataset~\cite{dataset_nyu} as well as the large-scale indoor dataset Occ-ScanNet~\cite{ssc_ISO}. The proposed approach is compared against state-of-the-art methods, and both experimental results and ablation studies are thoroughly analyzed. In addition, we present qualitative visualizations to further illustrate the effectiveness of our proposed method.

\definecolor{ceiling}{RGB}{214, 38, 40}
\definecolor{floor}{RGB}{43, 160, 43}
\definecolor{wall}{RGB}{158, 216, 229}
\definecolor{window}{RGB}{114, 158, 206}
\definecolor{chair}{RGB}{204, 204, 91}
\definecolor{bed}{RGB}{255, 186, 119}
\definecolor{sofa}{RGB}{147, 102, 188}
\definecolor{table}{RGB}{30, 119, 181}
\definecolor{tvs}{RGB}{188, 188, 33}
\definecolor{furniture}{RGB}{255, 127, 12}
\definecolor{objects}{RGB}{196, 175, 214}

\begin{table*}
    \centering
    \captionsetup{justification=raggedright,singlelinecheck=false} 
    \caption{\textbf{Comparisons among state-of-the-art MSSC methods} on the \textbf{Occ-ScanNet} dataset \cite{ssc_ISO}. The symbol $\dag$ denotes the results presented by~\cite{ssc_outdoor_MonoScene}. The \hlorange{Best} and \hlteal{2nd Best} scores from each metric are highlighted in \hlorange{Orange} and \hlteal{Teal}, respectively. Symbol * denotes the results obtained using their official code trained on the Occ-ScanNet dataset.}
    \vspace{-0.3cm}
    \resizebox{\linewidth}{!}{
    \begin{tabular}{r|r|c|ccccccccccc|c}
    \toprule
    Method & Venue & IoU (\%) & 
    \rotatebox{90}{\textcolor{ceiling}{$\blacksquare$} ceiling} & 
    \rotatebox{90}{\textcolor{floor}{$\blacksquare$} floor} & 
    \rotatebox{90}{\textcolor{wall}{$\blacksquare$} wall} & 
    \rotatebox{90}{\textcolor{window}{$\blacksquare$} window} & 
    \rotatebox{90}{\textcolor{chair}{$\blacksquare$} chair} & 
    \rotatebox{90}{\textcolor{bed}{$\blacksquare$} bed} & 
    \rotatebox{90}{\textcolor{sofa}{$\blacksquare$} sofa} & 
    \rotatebox{90}{\textcolor{table}{$\blacksquare$} table} & 
    \rotatebox{90}{\textcolor{tvs}{$\blacksquare$} tvs} & 
    \rotatebox{90}{\textcolor{furniture}{$\blacksquare$} furniture} & 
    \rotatebox{90}{\textcolor{objects}{$\blacksquare$} objects} & 
    \rotatebox{90}{mIoU (\%)} 
    \\
    \midrule\midrule
    TPVFormer~\cite{ssc_outdoor_TPVFormer} & CVPR'23 & $33.39$ & $6.96$ & $32.97$ & $14.41$ & $9.10$ & $24.01$ & $41.49$ & \hlteal{$45.44$} & $28.61$ & $10.66$ & $35.37$ & $25.31$ & $24.94$ 
    \\
    GaussianFormer~\cite{ssc_outdoor_gaussianformer} & ECCV'24 & $40.91$ & \hlteal{$20.70$} & $42.00$ & $23.40$ & $17.40$ & $27.0$ & \hlteal{$44.30$} & $44.80$ & $32.70$ & $15.30$ & $36.70$ & $25.00$ & $29.93$ 
    \\
    MonoScene~\cite{ssc_outdoor_MonoScene} & CVPR'22 & $41.60$ & $15.17$ & $44.71$ & $22.41$ & $12.55$ & $26.11$ & $27.03$ & $35.91$ & $28.32$ & $6.57$ & $32.16$ & $19.84$ & $24.62$ 
    \\
    ISO~\cite{ssc_ISO} & ECCV'24 & $42.16$ & $19.88$ & $41.88$ & $22.37$ & $16.98$ & \hlteal{$29.09$} & $42.43$ & $42.00$ & $29.60$ & $10.62$ & $36.36$ & $24.61$ & $28.71$ 
    \\
    SurroundOCC~\cite{wei2023surroundocc} & ICCV'23 & \hlteal{$42.52$} & $18.90$ & \hlteal{$49.30$} & \hlteal{$24.80$} & \hlteal{$18.00$} & $26.80$ & $42.00$ & $44.10$ & \hlteal{$32.90$} & \hlteal{$18.60$} & \hlteal{$36.80$} & \hlteal{$26.90$} & \hlteal{$30.83$} 
    \\
    \midrule

     \textbf{\model} & \textbf{Ours} & \hlorange{$54.63$} & \hlorange{$37.66$} & \hlorange{$57.34$} & \hlorange{$40.51$} & \hlorange{$29.49$} & \hlorange{$43.29$} & \hlorange {$60.41$} & \hlorange {$63.08$} & \hlorange{$47.05$} & \hlorange{$29.63$} & \hlorange{$54.08$} & \hlorange{$36.15$} & \hlorange{$45.33$} 
    \\
    \bottomrule
    \end{tabular}
    }
    \label{tab:Occ-ScanNet}
    \vspace{-0.1cm}
\end{table*}

\subsection{Experimental Setup}

\noindent\textbf{Datasets.}
The NYUv2 dataset~\cite{dataset_nyu} consists of $1,449$ depth images captured with a Kinect sensor across diverse indoor scenes, split into $795$ for training and $654$ for testing. Following SSCNet~\cite{SSCNet}, we employ the 3D voxel annotations provided by \cite{ssc15} and adopt the category mappings defined in \cite{ssc21}.
The Occ-ScanNet dataset~\cite{ssc_ISO} serves as a large-scale benchmark for indoor 3D semantic scene completion. It contains $45,755$ training frames and $19,764$ validation frames, with scenes voxelized into grids of $60 \times 60 \times 36$. Both NYUv2 and Occ-ScanNet are annotated with the same 12 semantic classes: one for empty (free space) and eleven for typical indoor categories, including \textit{ceiling}, \textit{floor}, \textit{wall}, \textit{window}, \textit{chair}, \textit{bed}, \textit{sofa}, \textit{table}, \textit{television}, \textit{furniture}, and \textit{miscellaneous objects}.

\noindent\textbf{Evaluation Metrics.} For fair comparison, we follow the evaluation protocol of SSCNet~\cite{SSCNet}. In semantic scene completion (SSC), performance is measured using the intersection over union (IoU) between predicted and ground-truth voxel labels for each semantic class, and the overall accuracy is reported as the mean IoU (mIoU) across all classes. For scene completion (SC), voxels are classified into empty or occupied, and performance is evaluated using IoU.

\noindent\textbf{Implementation details.} We implement our approach using the PyTorch framework. The model is optimized with stochastic gradient descent (SGD), starting from an initial learning rate of $0.1$ for NYUv2 and $0.003$ for Occ-ScanNet, which is adjusted according to the polynomial learning rate policy. We train the network for 100 epochs. The output 3D feature maps have dimensions of $60 \times 36 \times 60$ for NYUv2 and $60 \times 60 \times 36$ for Occ-ScanNet. All experiments are conducted on an NVIDIA A100 GPU.

\begin{table*}[t]
\centering
\setlength{\tabcolsep}{3pt} 
\renewcommand{\arraystretch}{1.1} 

\begin{minipage}{0.32\linewidth}
\centering
\caption{Ablation study on the core components in the proposed framework.}
\vspace{-0.3cm}
\resizebox{\linewidth}{!}{
\begin{tabular}{l|c|c}
\toprule
\textbf{Method} & \textbf{SC-IoU} (\%) $\uparrow$ & \textbf{SSC-mIoU} (\%) $\uparrow$ \\
\midrule\midrule
Baseline & $47.43$ & $28.68$ \\
$+$ ASA & $50.11$ ($+2.68$) & $30.98$ ($+2.30$) \\
$+$ CRPE & $50.71$ ($+0.60$) & $31.66$ ($+0.68$) \\
$+$ CMF & $\mathbf{51.33}$ ($+0.62$) & $\mathbf{32.50}$ ($+0.84$) \\
\bottomrule
\end{tabular}}
\label{tab:component_module}
\end{minipage}
\hfill
\begin{minipage}{0.32\linewidth}
\centering
\caption{Ablation study on different Adaptive Serialized Attention configurations.}
\vspace{-0.3cm}
\renewcommand{\arraystretch}{1.32}
\resizebox{\linewidth}{!}{
\begin{tabular}{l|c|c}
\toprule
\textbf{Method} & \textbf{SC-IoU} (\%) $\uparrow$ & \textbf{SSC-mIoU} (\%) $\uparrow$ \\
\midrule\midrule
Baseline & $47.43$ & $28.68$ \\
+ VanillaShift & $49.76$ ($+2.33$) & $30.61$ ($+1.93$) \\
+ GumbelShift & $50.87$ ($+1.11$) & $31.96$ ($+1.35$) \\
+ AnnealedShift & $\mathbf{51.33}$ ($+0.46$) & $\mathbf{32.50}$ ($+0.54$) \\
\bottomrule
\end{tabular}}
\label{tab:adaptive_serialized_attention}
\end{minipage}
\hfill
\begin{minipage}{0.32\linewidth}
\centering
\caption{Ablation study on the Center-Relative Positional Encoding setups.}
\vspace{-0.3cm}
\renewcommand{\arraystretch}{1.15}
\resizebox{\linewidth}{!}{
\begin{tabular}{l|c|c}
\toprule
\textbf{Method} & \textbf{SC-IoU} (\%) $\uparrow$ & \textbf{SSC-mIoU} (\%) $\uparrow$ \\
\midrule\midrule
\textit{w/o} CRPE & $50.85$ & $32.06$ \\
\textit{w/o} PVM & $50.89$ & $32.11$ \\
\textit{w/o} RYP & $50.96$ & $32.23$ \\
\textit{with} CRPE & $\mathbf{51.33}$ & $\mathbf{32.50}$ \\
\bottomrule
\end{tabular}}
\label{tab:center_relative_positional_encoding}
\end{minipage}
\end{table*}
\begin{figure}[t]
    \centering
    \includegraphics[width=\linewidth]{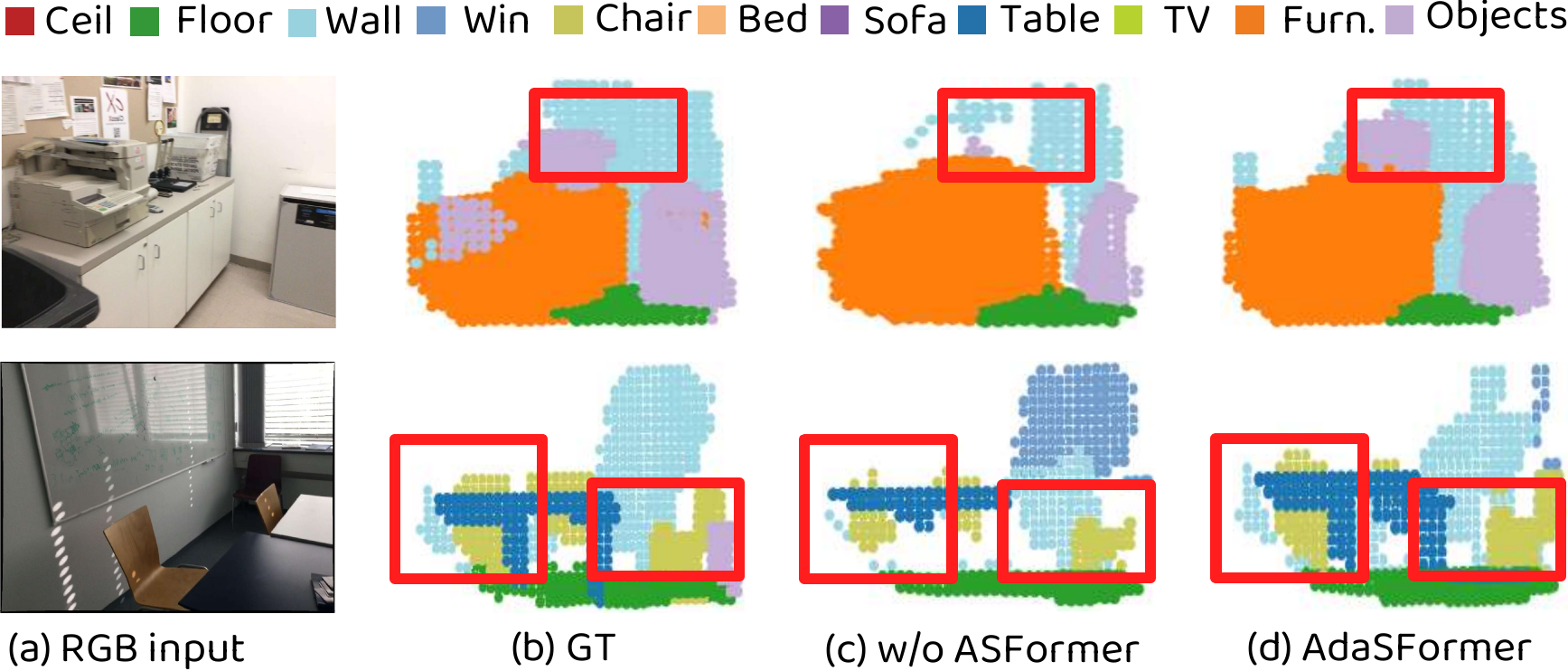}
    \vspace{-0.4cm}
    \caption{Visualization of the ablation results on the Occ-ScanNet dataset~\cite{ssc_ISO}. Introducing AdaSFormer significantly improves the model’s capability to capture spatial and semantic relationships, leading to more complete and coherent object completion.}
    \label{fig:occ_ablation}
    \vspace{-0.3cm}
\end{figure}

\subsection{Comparisons with State of the Arts}
\noindent\textbf{Quantitative Comparisons.} The results of the quantitative comparisons are presented in Tables ~\ref{tab:NYUv2} and ~\ref{tab:Occ-ScanNet}. It can be seen that our method achieves the best performance on the SSC task. On the NYU v2 dataset, our approach surpasses the current state-of-the-art method, ISO, by $1.25$ mIoU, achieving a total of $32.50$ mIoU. Similarly, on the OCC-ScanNet dataset, our method outperforms the best existing method, ISO, by $14.50$ mIoU, reaching $45.33$ mIoU.

\begin{figure*}[t]
    \centering
    \includegraphics[width=\linewidth]{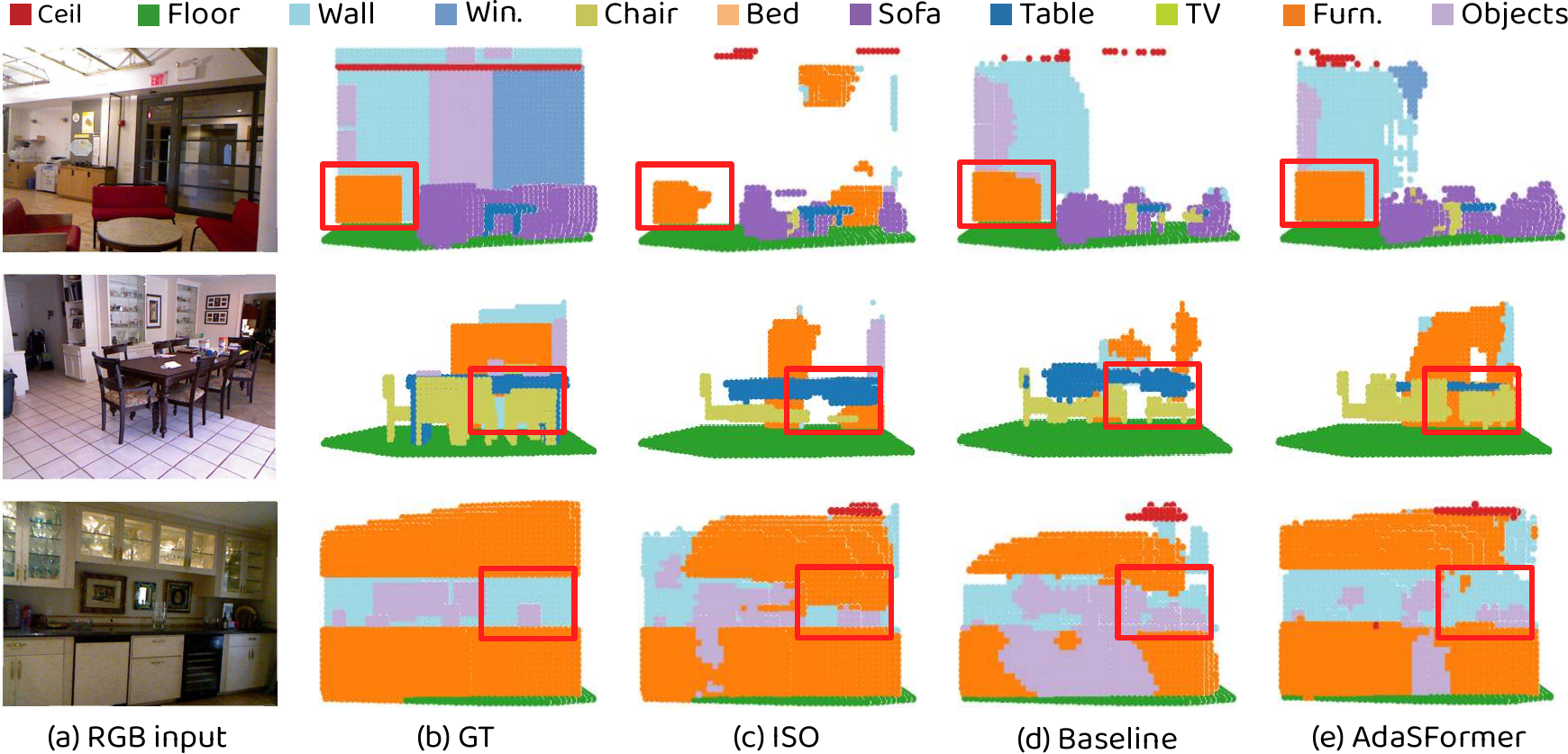}
    \vspace{-0.45cm}
    \caption{\textbf{Qualitative Comparisons} of semantic scene completion results on the NYUv2 testset~\cite{dataset_nyu} with different methods. Figures from left to right: (a) The single RGB input; (b) The ground truth; (c) ISO~\cite{ssc_ISO}; (d) Baseline; and (e) Our proposed method. } 
    \label{fig:nyu_vis}
    \vspace{-0.14cm}
\end{figure*}

\noindent\textbf{Qualitative Assessments.} The \cref{fig:nyu_vis} presents the visual comparison results of our method, the Baseline, and the ISO method on the NYU v2 dataset. It can be observed that our approach achieves superior performance in the Semantic Scene Completion (SSC) task. Benefiting from the proposed AdaSFormer model, our method is able to obtain a sufficiently large receptive field in the early stages of the network, enabling more effective capture of global spatial structures and enhancing the completeness and semantic consistency of the overall reconstruction. As shown in \cref{fig:nyu_vis}, it is evident that the increased receptive field facilitates more accurate scene completion and enhances the ability to distinguish between different objects in the scene.

\subsection{Ablation Study}
We present exhaustive ablation experiments on the NYUv2 dataset~\cite{dataset_nyu} to evaluate the effectiveness of our method from several perspectives, focusing on key components such as Adaptive Serialized Attention (ASA) and Center-Relative Positional Encoding (CRPE). 
Due to space constraints, more details and visualizations are provided in the supplementary material.

\noindent\textbf{The Effectiveness of Core Components.} 
Firstly, we conduct a systematic ablation study on the core components of proposed AdaSFormer, including the Adaptive Serialized Transformer (ASA), Center-Relative Positional Encoding (CRPE), and Conv-Modulated Former Block (CMF). 
By removing each module individually, we quantitatively analyze the contribution of each component to the overall performance, thereby gaining a deeper understanding of the rationale and advantages behind our design. 
The experimental results are presented in the \cref{tab:component_module}, where ASA, CRPE, and CMF denote the corresponding modules, and Baseline refers to the network with all three modules removed, retaining only the remaining convolutional structure. By incorporating ASA, we can observe an improvement over the baseline of $2.68$ in IoU and $2.30$ in mIoU. Adding CRPE further increases the performance by $0.60$ in IoU and $0.68$ in mIoU, while the inclusion of CMF brings an additional improvement of $0.62$ in IoU and $0.84$ in mIoU.

Furthermore, we present visualization results comparing the model with and without the AdaSFormer module to more intuitively demonstrate its effectiveness. 
As shown in \cref{fig:occ_ablation}, incorporating the AdaSFormer module significantly enhances the model’s spatial perception capability, enabling it to better capture the spatial layout and semantic relationships among objects in the scene. 
Consequently, the model can more effectively recover occluded or missing structures in complex environments.

\noindent\textbf{Design Choices of Adaptive Serialized Attention.} 
We also perform ablation experiments on several key design choices of the Adaptive Serialized Attention, including Vanilla Serialized Transformer, Gumbel-Softmax learning, and the temperature annealing strategy. As illustrated in \cref{tab:adaptive_serialized_attention}, `Baseline' represents the model without the Adaptive Serialized Transformer block. `VanillaShift' uses \texttt{torch.nn.Parameter} to directly learn $n$ discrete shifts for $n$ serialized intervals. `GumbelShift' adopts the straight-through Gumbel-Softmax for discrete shift learning. In addition, the temperature annealing strategy is introduced to progressively adjust the Gumbel-Softmax during training. We observe that introducing Gumbel-Softmax improves the SSC performance by $1.35$ mIoU, while further applying temperature annealing brings an additional $0.54$ mIoU gain, demonstrating the effectiveness of these strategies in stabilizing optimization and enhancing adaptive shift learning.

\noindent\textbf{Design Choices of Center-Relative Positional Encoding.} 
We further validate two key design choices of the center-relative positional encoding, including projected voxel mean (PVM) and the effectiveness of relative yaw and pitch positional encoding (RPY). As shown in \cref{tab:center_relative_positional_encoding}, for NYU v2, we observe that replacing the projected voxel mean (PVM) with the scene physical center at position $[30, 18, 30]$ results in a drop of $0.39$ mIoU, while removing the relative yaw and pitch positional encoding (RYP) leads to a decrease of $0.27$ mIoU.

\begin{table}[t]
    \centering
  
    \caption{Latency (in $\mathrm{ms}$) study on each component.}
    \vspace{-0.3cm}
    \resizebox{\linewidth}{!}{
    \begin{tabular}{l|c|c|c|c|c}
        \toprule
        \textbf{Method} &\textbf{ISO~\cite{ssc_ISO}}& \textbf{AdaSFormer} & \textbf{w/o ADA} & \textbf{w/o CRPE} & \textbf{w/o CMLN} \\
        \midrule\midrule
        Latency & $535.2$ & $178.9$ & $\mathbf{156.0}$ & $177.4$ & $174.3$ \\
        \bottomrule
    \end{tabular}}
    \label{tab:speed}
    \vspace{-0.4cm}
\end{table}

\noindent\textbf{Effect of Key Components on Computational Efficiency.} 

In the end, we evaluate the computational efficiency of each component, including Adaptive Serialized Attention (ASA), Center-Relative Positional Encoding (CRPE), and Convolution-Modulated Layer Normalization (CMLN).
As shown in \cref{tab:speed}, AdaSFormer achieves a latency of $178.9$ ms, which is significantly lower than ISO~\cite{ssc_ISO} ($535.2$ ms), demonstrating the efficiency of the proposed architecture.
Removing individual modules leads to only marginal latency variations ($156.0$ ms without ASA, $177.4$ ms without CRPE, and $174.3$ ms without CMLN), indicating that all components are lightweight.
ASA introduces a small overhead of $22.9$ ms while contributing substantially to performance improvement, confirming the balance between efficiency and accuracy.

\section{Conclusion}

In this paper, we propose a novel Adaptive Serialized Transformer (AdaSFormer) framework for indoor monocular semantic scene completion, which incorporates three key designs: 1) an Adaptive Serialized Attention that enables a larger, high-resolution, and adaptive receptive field; 2) a Center-Relative Positional Encoding that captures the spatial information richness to better represent scene structure; and 3) a Conv-Modulated Layer Normalization that facilitates the integration of Transformer and convolutional architectures, leading to a lightweight yet effective design. Extensive experiments on NYUv2 and Occ-ScanNet demonstrate the superiority of our method, achieving state-of-the-art performance across multiple datasets.

\clearpage\clearpage
\section*{Acknowledgments}
This work was supported by the National Natural Science Foundation of China under Grant No.\ 62071330.

The authors would like to sincerely thank the Program Chairs, Area Chairs, and Reviewers for the time and effort devoted during the review process.
\vspace{0.1cm}

{
    \small
    \bibliographystyle{ieeenat_fullname}
    \bibliography{main}
}

\end{document}